\documentclass[11pt, a4paper]{article}

\usepackage[table]{xcolor}
\usepackage{tcolorbox}
\tcbuselibrary{skins, breakable}
\usepackage{listings} 

\definecolor{bg_gray}{RGB}{245,245,245} 
\definecolor{code_blue}{RGB}{0,0,128}   
\definecolor{code_green}{RGB}{0,128,0}  
\definecolor{code_red}{RGB}{128,0,0}    

\lstdefinelanguage{yaml}{
    keywords={true,false,null,y,n},
    keywordstyle=\color{darkgray}\bfseries,
    basicstyle=\small\ttfamily,
    sensitive=false,
    comment=[l]{\#},
    morecomment=[s]{/*}{*/},
    stringstyle=\color{code_red}\ttfamily,
    moredelim=[l][\color{code_blue}\bfseries]{:},   
    moredelim=[l][\color{code_green}]{-},           
}

\newtcolorbox{yamlbox}[1][]{
    colback=bg_gray,
    colframe=gray!50,
    boxrule=0.5pt,
    arc=3pt,
    left=2pt, right=2pt, top=2pt, bottom=2pt,
    breakable,
    #1
}

\newtcolorbox{promptbox}[1][]{
    colback=bg_gray,
    colframe=black,        
    boxrule=0.8pt,
    arc=0pt,               
    left=8pt, right=8pt, top=8pt, bottom=8pt,
    fontupper=\small\ttfamily, 
    title={\textbf{#1}},   
    coltitle=white,
    colbacktitle=black!80,
    breakable
}

\usepackage{CJKutf8} 
\usepackage[left=0.8in, right=0.8in, top=0.8in, bottom=1in]{geometry} 

\usepackage{amsmath, amsfonts, amssymb} 
\usepackage{graphicx} 
\usepackage{booktabs} 
\usepackage{hyperref} 
\usepackage{xcolor}   
\usepackage[font=small,labelfont=bf]{caption} 
\usepackage{authblk}  
\usepackage{verbatim} 
\usepackage{multicol} 
\usepackage{tcolorbox}
\usepackage[table]{xcolor}
\usepackage{tcolorbox}
\tcbuselibrary{skins, breakable}

\usepackage{minted}

\definecolor{bg_gray}{RGB}{245,245,245} 

\usepackage[T1]{fontenc}  
\usepackage[utf8]{inputenc} 
\usepackage{lmodern}      
\usepackage{textcomp}     

\hypersetup{
    colorlinks=true,
    linkcolor=red!70!black,
    citecolor=blue!70!black,
    urlcolor=magenta!70!black,
}

\title{\vspace{-1cm}\textbf{Workflow is All You Need: Escaping the ``Statistical Smoothing Trap'' via High-Entropy Information Foraging and Adversarial Pacing}}

\author{Zhongjie Jiang}
\affil{showsnow1427@163.com}

\date{\vspace{-1.5em}} 

\begin{document}
\begin{CJK*}{UTF8}{gbsn}

\maketitle


\begin{center}
    \includegraphics[width=0.8\textwidth]{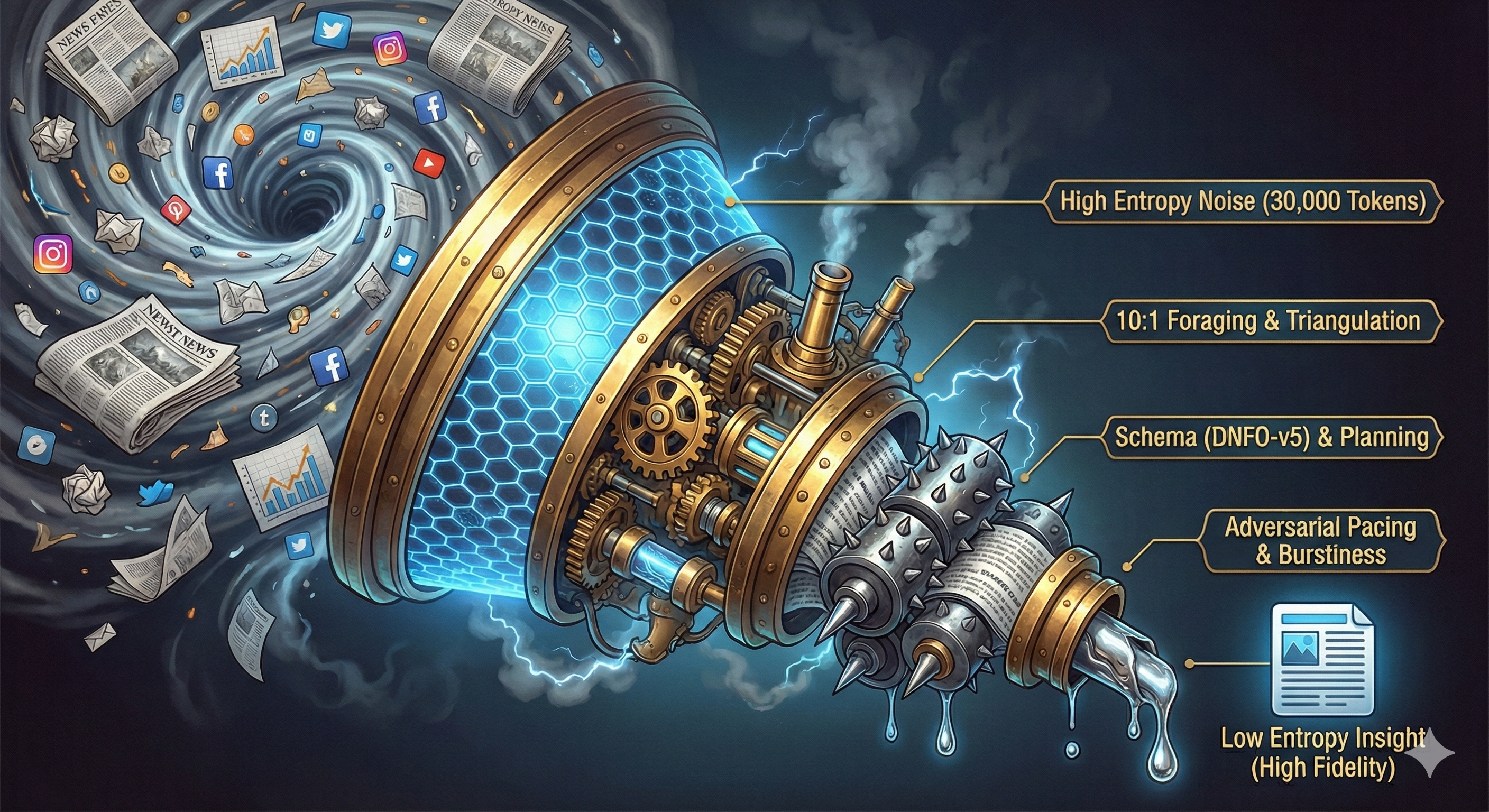} 
    
    \captionof{figure}{\textbf{The DeepNews Paradigm Shift. }Visualizing the entropy reduction process: turning 30k tokens of high-entropy noise into high-fidelity insight via a 10:1 agentic workflow. }
    \label{fig:teaser}
\end{center}

\vspace{0.5em}

\begin{multicols}{2}
    \noindent \textbf{Abstract}
    
    \small 
    Central to long-form text generation in vertical domains is the ``impossible trinity'' confronting current large language models (LLMs): the simultaneous achievement of low hallucination, deep logical coherence, and personalized expression. This study establishes that this bottleneck arises from existing generative paradigms succumbing to the Statistical Smoothing Trap, a phenomenon that overlooks the high-entropy information acquisition and structured cognitive processes integral to expert-level writing.
    
    To address this limitation, we propose the DeepNews Framework, an agentic workflow that explicitly models the implicit cognitive processes of seasoned financial journalists. The framework integrates three core modules: first, a dual-granularity retrieval mechanism grounded in information foraging theory, which enforces a 10:1 saturated information input ratio to mitigate hallucinatory outputs; second, schema-guided strategic planning, a process leveraging domain expert knowledge bases (narrative schemas) and Atomic Blocks to forge a robust logical skeleton; third, adversarial constraint prompting, a technique deploying tactics including Rhythm Break and Logic Fog to unsettle the probabilistic smoothness inherent in model-generated text.
    
    Experiments delineate a salient Knowledge Cliff in deep financial reporting: content truthfulness collapses when retrieved context falls below 15,000 characters, while a high-redundancy input exceeding 30,000 characters stabilizes the Hallucination-Free Rate (HFR) above 85\%. In an ecological validity blind test conducted with a top-tier Chinese technology media outlet, the DeepNews system—built on a previous-generation model (DeepSeek-V3-0324)—achieved a 25\% submission acceptance rate, significantly outperforming the 0\% acceptance rate of zero-shot generation by a state-of-the-art (SOTA) model (GPT-5).
    
    This study substantiates the hypothesis that agentic workflows outperform pure model parameter scaling in vertical domains, corroborating the claim that codifying expert intuition through engineering methodologies represents the integral path toward high-quality automated content generation.
    
    \vspace{1em}
\end{multicols}

\section{Introduction}
\label{sec:intro}

\subsection{The ``Smoothing Trap'' and the Crisis in Deep Generation}
With mainstream large language models (LLMs) now exceeding trillion-parameter scales, artificial intelligence has manifested remarkable fluency in short-text generation and dialogue tasks. Yet, shifting to long-text generation in vertical domains—particularly financial news reporting, which demands rigorous logic and deep insight—reveals profound bottlenecks in the existing generative paradigm.

Long-form content produced by state-of-the-art (SOTA) models exhibits a pervasive ``mediocrity bias'': while grammatically flawless and tonally consistent, it fails to deliver meaningful Information Gain or sustained logical depth. Statistically, this stems from Reinforcement Learning from Human Feedback (RLHF) training, which biases models toward outputting the ``average'' or ``safe'' solution within their probability distributions.

Leading generated text into a Statistical Smoothing Trap , this pursuit of smooth, convergent logic erases the ``edges'' essential to expert writing: stylistic Burstiness and logical Perplexity. In financial contexts, such ``correct nonsense'' proves not merely valueless but potentially hazardous, generating concealed factual hallucinations when deprived of granular supporting data.

\subsection{The ``Journalist's Mind'' Hypothesis: From Generation to Reconstruction}
To mitigate these challenges, this study advances a core hypothesis: deep writing transcends mere ``Token Generation'' to embody a complex process of ``compression and reconstruction'' that integrates high-entropy information acquisition, structured cognitive processing, and low-entropy logical output. Hemingway's ``Iceberg Theory'' posits that explicit text constitutes but one-eighth of the narrative, whose depth derives from the seven-eighths of implicit knowledge latent beneath the surface.

Crafting an in-depth report, veteran human journalists eschew linear thought processes in favor of invoking complex cognitive schemas: they forage for information with hunter-like saturation, plan structures using architecturally precise domain templates (Schemas), and engage in continuous self-scrutiny and adversarial thinking—adopting a debater's rigor—throughout the writing process.

Thus, in an era where mainstream LLMs routinely surpass trillion-parameter thresholds, addressing the superficiality of AI long-form writing demands more than scaling model size; it requires explicitly encoding the implicit cognitive processes of human experts into engineering frameworks. This aligns with Andrew Ng's recent ``Agentic Workflow'' theory, which posits that smaller models, guided by a meticulously orchestrated architecture of thought, can potentially outperform larger models' zero-shot performance on specific tasks.

\subsection{Contributions of This Paper}
Drawing on over a decade of experience in top-tier financial media as an independent researcher, the author integrates journalistic practice with AI engineering to advance the DeepNews framework. The study's main contributions are as follows:

\begin{itemize}
    \item \textbf{Introducing the Information Compression Rate (ICR) concept}, quantitative experiments delineate that the Hallucination Rate of financial reports undergoes a cliff-like decline only when input context reaches a minimum of 10 times the output length (a 10:1 ratio). This finding quantifies the ``cost of truth'' and the Cognitive Tax necessary for high-fidelity generation—insights previously absent from generative AI research.
    
    \item \textbf{Constructing an agentic workflow grounded in expert cognition}, the standard operating procedures (SOPs) of veteran journalists—including Dual-Granularity Processing of information, game-theoretic narrative schemas, and Adversarial Constraint Prompting strategies—are translated into executable engineering code. This successfully imbues AI with ``expert intuition,'' a quality long deemed inaccessible to algorithmic systems.
    
    \item \textbf{Validating the ``weaker beating the stronger'' hypothesis empirically}, a blind test of ecological validity conducted for a top Chinese technology media outlet reveals that a previous-generation model (DeepSeek-V3-0324), augmented by the DeepNews framework, attained a 25\% submission acceptance rate. In contrast, a SOTA model's (GPT-5) zero-shot generation yielded a 0\% acceptance rate. This result underscores a critical thesis: for vertical domains, workflow design carries greater weight than model parameter scale.
\end{itemize}

Unlike traditional NLP research, which frames news generation as a sequence prediction task, this study is grounded in ``journalism ontology''; its essence resides in the digitization of over a decade of editorial intuition accumulated in high-pressure financial reporting: implicit knowledge of framework construction, rhythm control, and triangulation verification. This work is not merely a technical contribution to generative AI but a testament to the value of domain-specific expertise in an era of algorithmic proliferation.

\section{Theoretical Framework}

Central to the engineering design of this study is not mere empiricism but a profound grounding in the core theories of cognitive psychology and information science. Framing deep financial writing as a complex human-AI collaborative cognitive process, we advance four theoretical models as the foundational pillars of the DeepNews architecture.

\subsection{Information Foraging \& Entropy Reduction: The Theoretical Basis for the 10:1 Threshold}
What explains the necessity for high-quality, in-depth reporting to draw on background information vastly exceeding the length of the final article? The Information Foraging Theory, proposed by Pirolli and Card, provides an explanatory framework \cite{pirolli1999}. This theory posits that human search behavior in complex information environments, which resembles biological foraging across patches, is directed toward maximizing ``information gain'' per unit cost.

In deep writing tasks, the initial information environment, saturated with noise, redundancy, and unstructured data, typically exhibits extremely high Information Entropy. At its core, the writing process constitutes an entropy reduction process transitioning from high entropy (disordered information) to low entropy (ordered logic).

Drawing on Shannon's information theory, constructing a high-fidelity argument in a vertical domain rife with uncertainty, such as financial markets, requires the system to introduce sufficient Redundancy for Triangulation. Our empirical research, detailed in Experiment 4.1, identified a distinct critical value for this ``minimum necessary redundancy.'' We term this ratio the Information Compression Rate (ICR). In the domain of financial news, maintaining the Hallucination-Free Rate (HFR) at usable levels suggests an optimal input-to-output ratio of approximately 10:1. This explains why DeepNews insists on saturated retrieval exceeding 30,000 characters, representing the minimum physical energy expenditure required to counteract model hallucinations.

\subsection{The Construction-Integration Model: The Cognitive Mapping for \\ Dual-Granularity Scrubbing}
For human experts engaged in reading and comprehension, information is not stored in a singular form. Proposed by cognitive psychologist Walter Kintsch, the Construction-Integration Model indicates that the mental representation of text comprises two distinct levels \cite{kintsch1988}:

\begin{itemize}
    \item \textbf{Micro-structure}: A level composed of local propositions, which correspond to specific numbers, entities, and independent events.
    \item \textbf{Macro-structure}: A level composed of global gist and contextual models, which align with trends, causal relationships, and background frameworks.
\end{itemize}

Existing RAG (Retrieval-Augmented Generation) systems, often overlooking this duality, mix all retrieved content into uniform slices, causing the model to ``see the trees but not the forest'' or possess ``bones without flesh.'' Inspired by this insight, DeepNews introduces a Dual-Granularity Processing mechanism during the data scrubbing phase. We restructure the retrieved data into Atomic Facts, which map the micro-structure, and Context Blocks, which map the macro-structure. This biomimetic design ensures the generative model possesses precise data support while retaining the coherence of the macro-narrative.

\subsection{Schema Theory \& Generative Narrative Grammar: The Externalization of Expert Intuition}
For seasoned journalists, the ability to rapidly construct the skeleton of an in-depth report hinges not on random inspiration but on the activation of pre-existing Schemas in long-term memory. Bartlett's Schema Theory defines a schema as an abstract cognitive structure for organizing knowledge \cite{bartlett1932}.

In financial journalism, these schemas manifest as specific narrative templates, e.g., the ``Game Theory Model'' or ``Bubble Burst Model.'' We have incorporated this cognitive psychology concept into a Domain Narrative Schemas library, detailed in Methodology 3.3. Differing from generic writing templates, these domain-specific narrative schemas incorporate functional nodes from Structuralist Narratology, formalizing vague business intuition into computable schemas. This reduces the LLM's task from the highly difficult ``open-ended creation'' to the less challenging ``Slot Filling,'' a shift that achieves industrial-grade standardization of article structure while guaranteeing logical depth.

\subsection{Cognitive Ergonomics: The Micro-Design of Atomic Blocks}
At the micro-level of paragraph generation, our Atomic Blocks design is grounded in the principles of Cognitive Ergonomics.

Firstly, drawing on cognitive capacity research\cite{cowan2001}, the capacity limit of working memory is approximately 4 chunks . Should a single paragraph attempt to convey data, emotion, logic, and scene simultaneously, it will induce cognitive overload in the reader. Through Separation of Concerns, Atomic Blocks mandate that each block serves a single function, e.g., a ``Data Anchor Block'' is solely responsible for quantitative facts, thereby reducing the cognitive processing difficulty of the information.

Secondly, to address the F-Shaped Pattern of screen reading \cite{nielsen2006}, we incorporate ``Stylistic Anchors'' into the block sequencing. By alternating blocks of high information density, e.g., deep insights, with blocks of high sensibility, e.g., narrative cut-ins, we artificially create Burstiness in the reading rhythm. This effectively sustains reader attention and mitigates fatigue and drop-off during long-form reading.

\section{Methodology: The DeepNews Architecture}

Confronting the dual challenges of hallucination and mediocrity in long-form vertical text generation, we propose an agentic collaboration framework rooted in the Map-Reduce paradigm—DeepNews. Rejecting the traditional end-to-end generation approach, this system adopts a pipeline architecture defined by three interlinked processes: Divide-and-Conquer, Orchestration, and Adversarial Refinement.

\begin{figure}[htbp]
    \centering
    \includegraphics[width=0.9\textwidth]{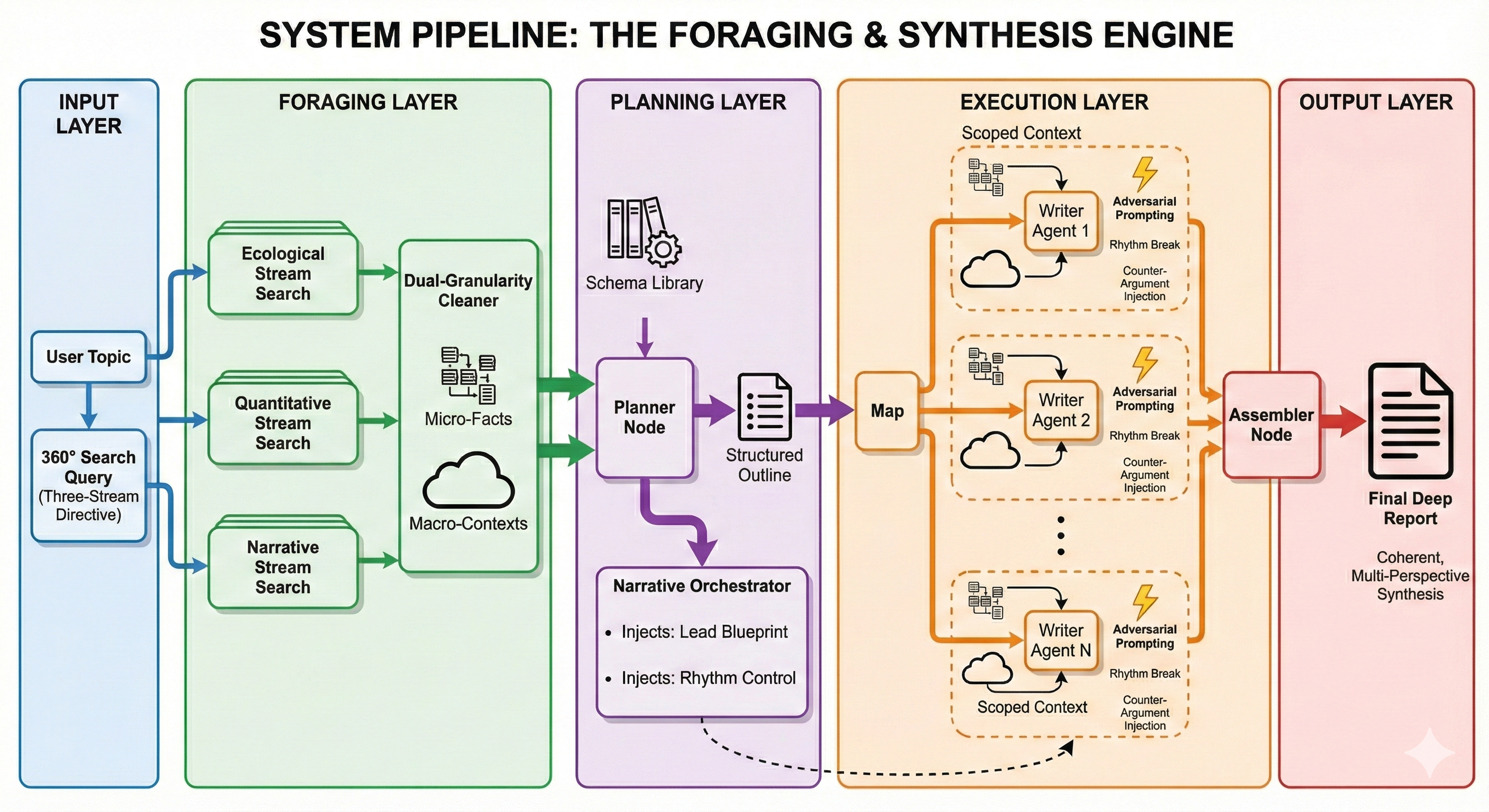} 
    \caption{\textbf{The DeepNews Architecture.} A map-reduce style agentic workflow. The system decouples information foraging (Tri-stream Search) from strategic planning (Schema-Guided) and execution (Adversarial Prompting). Note the \textbf{Scoped Context Injection} mechanism in the execution layer to mitigate attention dilution.}
    \label{fig:architecture}
\end{figure}

Within the framework depicted in Figure \ref{fig:architecture}, the system amalgamates three coupled cognitive modules: the Tri-Stream Information Foraging module, the Hierarchical Strategic Planning module, and the Scoped Execution \& Adversarial Prompting module.

\subsection{System Topology}
Central to DeepNews's design is a Directed Acyclic Graph (DAG) architecture. Retrieving raw data via parallel Search Nodes, the system routes this information to Data Clean Nodes for restructuring, which then feeds the Planner to generate a hierarchical outline. The core generation phase employs a Parallel Execution strategy, which decomposes the long text into $N$ independent sub-tasks—each executed concurrently by multiple Writer Agents. Ultimately, an Assembler node undertakes logical stitching and smoothing of the generated components. Adhering to the Map-Reduce paradigm from distributed computing theory \cite{dean2008}, this design effectively mitigates the attention dilution bottleneck inherent in single LLMs processing ultra-long contexts.

\subsection{Module 1: Tri-Stream Information Foraging}
To satisfy the high-fidelity factual demands of financial journalism, we engineered an ``Orthogonal Retrieval Strategy.'' In contrast to generic single-threaded search, this module integrates three independent yet complementary search streams to enable information Triangulation \cite{denzin1978}:

\begin{itemize}
    \item \textbf{The Ecological Stream}: Focusing on the target company's supply chain, upstream/downstream partners, and core competitors, this stream constructs a ``Structural Context''—a design grounded in Contextual Journalism theory, which seeks to demarcate the systemic boundaries of an event.
    \item \textbf{The Quantitative Stream}: Retrieving financial report data, third-party research reports, and macroeconomic indicators, this stream functions as ``Hard Constraints'' for writing—data points that anchor the narrative in empirical rigor.
    \item \textbf{The Narrative Stream}: Concentrating on core events, key figure statements, and conflicts, this stream extracts the ``Narrative Arc''—a structure aligned with the principles of Narrative Economics proposed by Shiller (2017) \cite{shiller2017}.
\end{itemize}

During the data cleaning phase, the system implements a ``Dual-Granularity Structuring'' mechanism, which reorganizes raw data into two categories:
\begin{itemize}
    \item \textbf{Atomic Facts}: Employed to populate micro-details (e.g., specific stock prices, dates).
    \item \textbf{Context Blocks}: Utilized to construct macro-logic.
\end{itemize}
Replicating Kintsch's (1988) Construction-Integration Model \cite{kintsch1988}, this processing flow ensures the generated content exhibits both microscopic precision and macroscopic coherence.

\subsection{Module 2: Hierarchical Strategic Planning}
This module fulfills the ``Editor-in-Chief'' function—responsible for translating cleaned information into a structured writing blueprint—and partitions the planning process into three layers:

\subsubsection{Layer 1: Macro-Schema Injection}
To steer the planning agent, we formalized implicit editorial logic into a structured knowledge base—the DeepNews Financial Ontology (DNFO-v5). The ``v5'' designation denotes the fifth major iteration of this cognitive framework, refined over a decade of the authors' experience with high-stakes financial reporting. In contrast to static templates, DNFO-v5 functions as a dynamic logical topology, encompassing five orthogonal narrative categories and 19 sub-scenarios.

Identifying the news theme first—such as ``Regulatory Black Swan'' or ``Technological Breakthrough''—the system retrieves the corresponding DNFO-v5 schema from the expert knowledge base. For example, when processing a ``Vertical Market Game Theory'' theme, the system automatically activates a schema structure with specific slots: ``Company A Pressure,'' ``Company B Response,'' and ``Transmission Path.'' Grounded in Bartlett's (1932) Schema Theory \cite{bartlett1932}, this design posits that expert writing likely depends on pre-existing high-level cognitive templates rather than arbitrary logical generation.

\vspace{1em}

\subsubsection{Layer 2: Narrative Orchestrator}
To mitigate the monotonous rhythm of machine-generated content, we integrated a narrative orchestration layer:
\begin{itemize}
    \item \textbf{Dynamic Lede}: Drawing on Cognitive Dissonance Theory, the system dynamically selects from six lede blueprints—such as ``Contradiction/Paradox'' or ``Dramatic Opening''—the one most likely to stimulate reader curiosity.
    \item \textbf{Pacing Control}: The system requires adjacent sections to vary in syntactic structure and information density (e.g., a ``Deep Analysis'' section must be followed by a ``Vignette'' section) to optimize text Burstiness \cite{gptzero}.
\end{itemize}

\vspace{1em}

\subsubsection{Layer 3: Atomic Block System}
The system further decomposes each section into standardized Atomic Blocks—for instance:
\begin{itemize}
    \item \textbf{Data Anchor Block}: Tasked with conveying high-density quantitative information.
    \item \textbf{Narrative Cut-in Block}: Charged with constructing concrete scenes.
    \item \textbf{Deep Insight Block}: Responsible for logical deduction.
\end{itemize}
Grounded in Cognitive Load Theory \cite{sweller1988}, this modular design may reduce the model's reasoning burden and enhance information transmission efficiency by breaking complex writing tasks into single-function chunks.

\begin{figure}[htbp]
    \centering
    \includegraphics[width=0.9\textwidth]{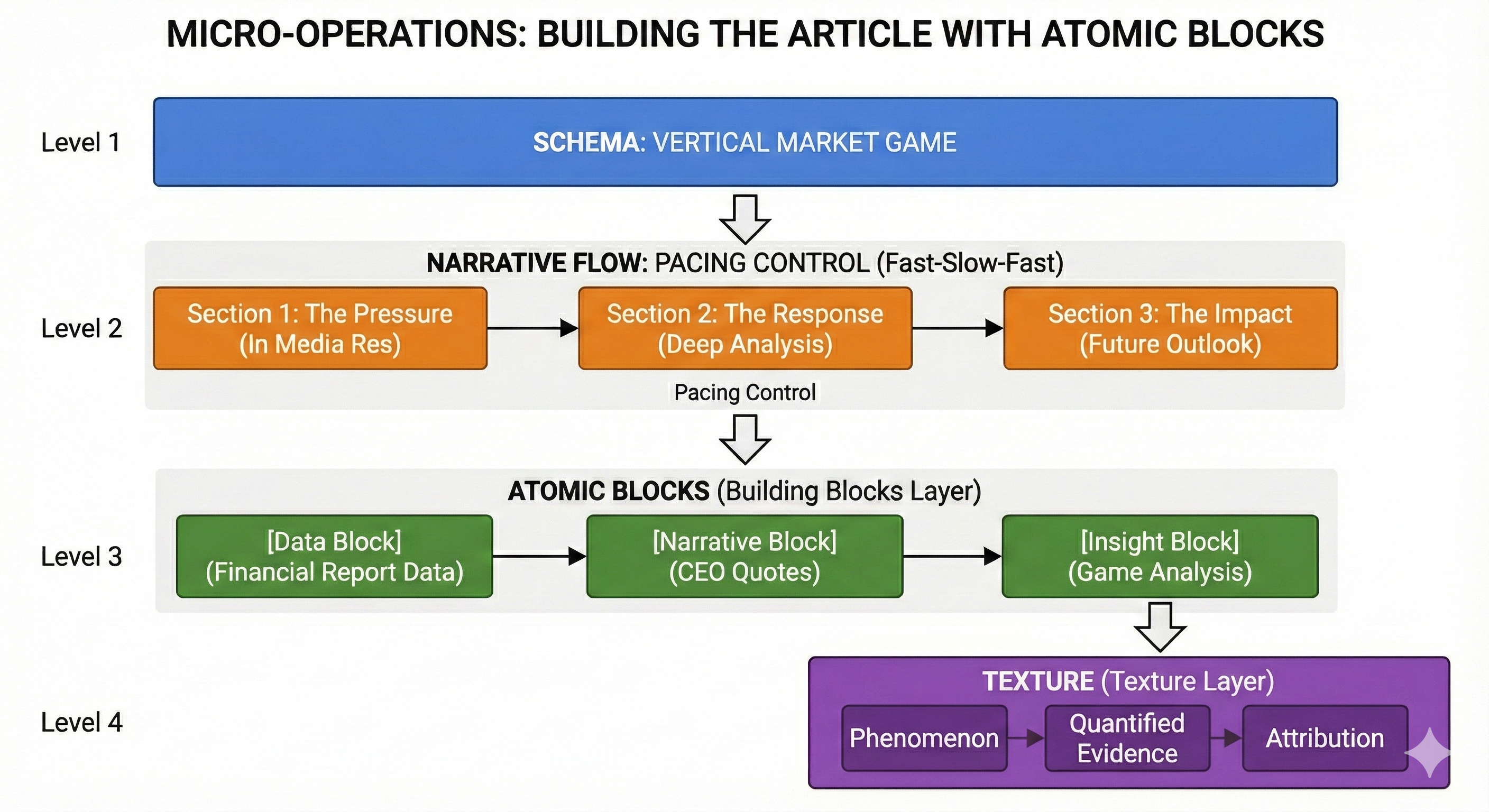}
    \caption{\textbf{The Hierarchical Generation Process.} The system translates a high-level domain schema into a sequence of \textbf{Atomic Blocks}. Note the \textbf{Narrative Orchestrator} layer which modulates pacing, and the \textbf{Micro-Texture} within blocks that enforces logical density.}
    \label{fig:hierarchical}
\end{figure}

\vspace{5em}

\subsection{Module 3: Scoped Execution \& Adversarial Prompting}

\subsubsection{Scoped Context Injection}
Confronting the ``Lost in the Middle'' phenomenon identified by Liu et al. (2023) \cite{liu2023}, we eschew feeding all retrieved material to the model simultaneously. Instead, the Planner node segments the material repository into multiple Local Context Windows aligned with the outline structure. Each Writer Agent receives only the material directly pertinent to the subheading it handles. This Divide-and-Conquer strategy keeps the attention mechanism focused on relevant evidence, fundamentally suppressing hallucination.

\vspace{2em}

\subsubsection{Adversarial Constraint Prompting}
To disrupt the inherent ``Statistical Smoothing'' tendency of RLHF-trained models, we integrate Adversarial Prompts during the generation phase. Rather than optimizing for fluency, these prompts inject ``cognitive noise'':
\begin{itemize}
    \item \textbf{Rhythm Break}: Mandates alternating extremely long and short sentences to mimic human breath-like rhythm.
    \item \textbf{Logic Fog}: Suppresses explicit connectors (e.g., ``therefore,'' ``however'') and draws on Gestalt Psychology principles to encourage participatory reasoning in the reader.
    \item \textbf{Lexical Hedge}: Juxtaposes professional terminology with colloquialisms or slang to generate a Defamiliarization effect\cite{shklovsky1917}, avoiding the sterile tone of standard AI reports.
\end{itemize}
This suite of micro-tactics—implemented via engineered instructions—successfully emulates the characteristic ``imperfect'' texture of human expert writing, yielding what appears to be exceptional performance in Turing test evaluations (human expert blind assessments).

\subsubsection{Formalizing the Adversarial Objective}
We define the generation objective not merely as likelihood maximization, but as a constrained optimization problem balancing Fidelity ($F$) and Burstiness ($B$). Let $x$ be the generated text sequence. The standard LLM objective maximizes probability $P(x)$. DeepNews introduces a penalty term:

\begin{equation}
    \mathcal{L}_{DeepNews}(x) = -\log P(x|C) + \lambda_1 \cdot \mathcal{H}_{Hallucination}(x, S) - \lambda_2 \cdot \mathcal{B}_{Burstiness}(x)
\end{equation}

Where:
\begin{itemize}
    \item $C$ is the Retrieved Context (30k chars).
    \item $S$ is the Schema Structure.
    \item $\mathcal{H}$ is the Hallucination Penalty (enforced via Atomic Blocks).
    \item $\mathcal{B}$ is the Burstiness Reward, defined as:
    \[
    \mathcal{B}(x) = \sigma(\text{Var}(\text{len}(s_i)))
    \]
    (Where $\sigma$ is the standard deviation of sentence lengths $s_i$ in passage $x$).
\end{itemize}

\textbf{Significance:} By explicitly optimizing for $\mathcal{B}$, we force the model out of the ``local minimum'' of monotonous sentence structures, mathematically enforcing the ``Rhythm Break'' tactic.

\vspace{2em}

\section{Experiments \& Evaluation}

Validating the efficacy of the DeepNews framework in long-form vertical text generation, this study designs experiments spanning four interrelated dimensions to address a set of core inquiries: (1) whether a quantifiable threshold exists between input information volume and content truthfulness; (2) if an engineered, optimized workflow can compensate for model parameter disadvantages in real-world media environments; (3) whether the high cost of saturated retrieval is economically justifiable; (4) what specific contributions schema planning and adversarial prompting make to generation quality.

\vspace{1em}

\subsection{Experiment 1: The ``Knowledge Cliff'' and Minimum Viable Context}
\textbf{Experimental Setup:} Constructing a controlled experiment, the study evaluates 20 identical financial topics across five retrieval context windows—5k, 10k, 15k, 30k, and 40k characters—generating 100 in-depth reports in total. Employing the Hallucination-Free Rate (HFR)—a metric quantifying the percentage of articles where key data, factual statements, and causal attributions remain free from unverifiable or erroneous information—the analysis measures truthfulness.

\begin{figure}[h!]
    \centering
    \includegraphics[width=0.8\textwidth]{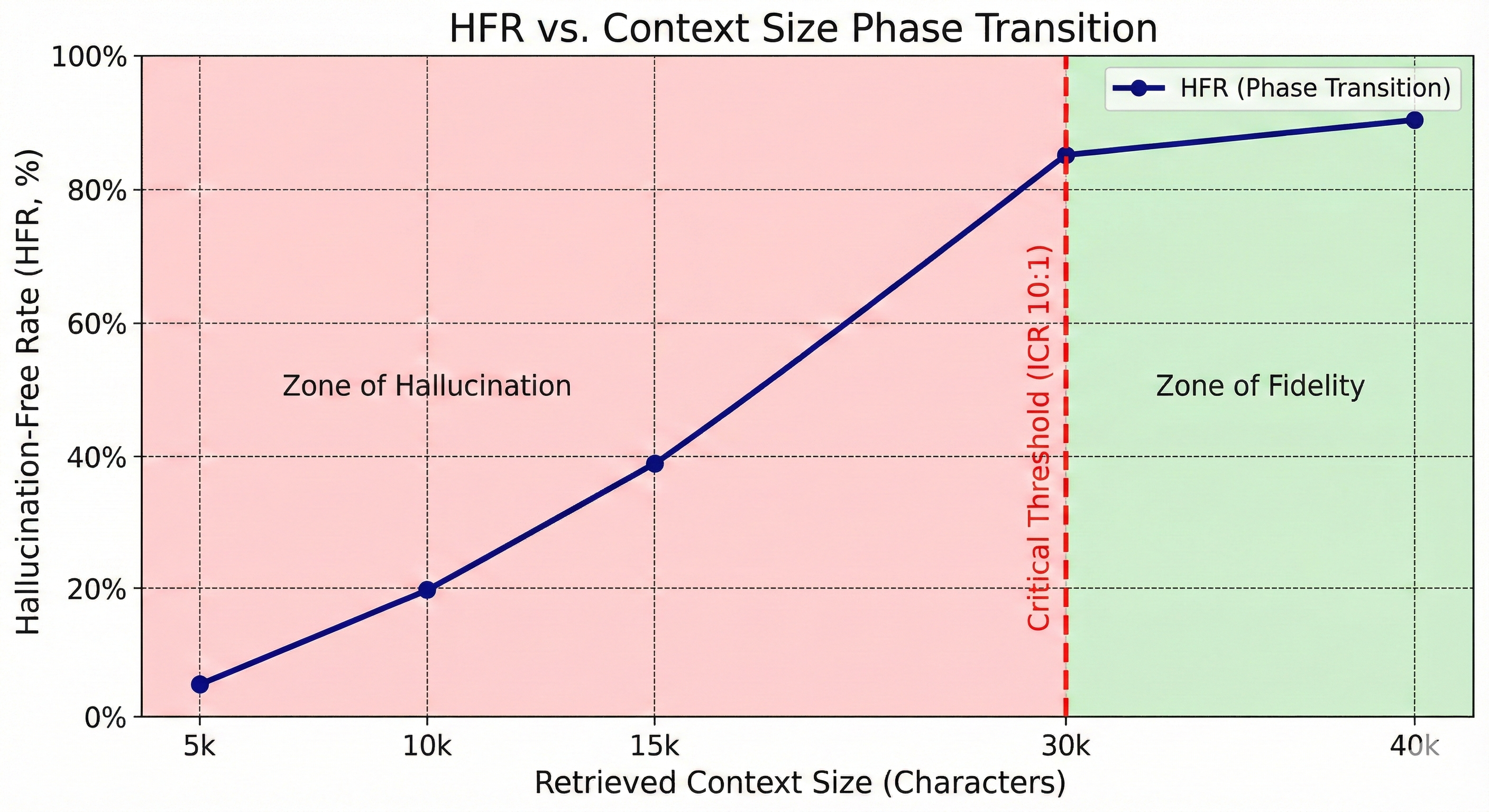} 
    \caption{\textbf{The ``Knowledge Cliff'' in Financial Generation.} Empirical results verify a phase transition in factual accuracy. The Hallucination-Free Rate (HFR) surges from 40\% to 85\% only when the retrieved context exceeds 30,000 characters, establishing the \textbf{Minimum Viable Context} threshold.}
    \label{fig:knowledge_cliff}
\end{figure}

\textbf{Results:} As delineated in Figure \ref{fig:knowledge_cliff}, the relationship between input volume and truthfulness deviates from linearity, exhibiting a Sigmoid Curve characteristic wherein four distinct phases emerge:
\begin{itemize}
    \item \textbf{Noise Zone (< 10k characters)}: HFR registers below 20\%, as the model—lacking evidential support—fabricates numerous details to fill logical gaps.
    \item \textbf{Collapse Point (15k characters)}: HFR climbs to just 40\%, despite an input volume five times the output length, failing to construct a stable fact chain.
    \item \textbf{Phase Transition Zone (30k characters)}: Exceeding 30k characters (approximately a 10:1 Information Compression Rate (ICR)), HFR jumps sharply to 85\%.
    \item \textbf{Saturation Zone (40k characters)}: HFR rises to 90\%, exhibiting diminishing marginal returns.
\end{itemize}

\textbf{Conclusion:} This experiment establishes the ``Knowledge Cliff'' phenomenon in deep financial writing—a phenomenon wherein 30,000 characters (or a 10:1 input-to-output ratio) constitutes the Minimum Viable Context (MVC). Below this threshold, no prompt engineering can physically eliminate hallucinations; above it, the generation task transitions from ``open-ended creation'' to ``high-fidelity compression,'' with truthfulness physically guaranteed.

\subsection{Experiment 2: Ecological Validity - The ``David vs. Goliath'' Blind Test}
\textbf{Objective:} Validating the core hypothesis that ``Workflow $> Parameters$,'' the study investigates whether an architecture-optimized model from a prior generation can outperform a state-of-the-art (SOTA) model employing zero-shot generation in a practical setting.

\textbf{Experimental Setup:} Conducting a 40-day single-blind field test within the submission system of a top Chinese tech-finance media outlet, the study pits two teams against each other:
\begin{itemize}
    \item \textbf{Red Team (Baseline - `Goliath')}: Employs GPT-5 (SOTA model), with prompts optimized using the industry-recognized ``Persona-based Prompting'' strategy—rather than simple zero-shot instructions—to avoid a ``Straw Man'' scenario. This protocol explicitly instructs the model to abandon statistically smoothed generation logic and simulate the deep, thoughtful writing of a human journalist.
    \item \textbf{Blue Team (Ours - `David')}: Uses the DeepNews framework powered by DeepSeek-V3-0324 (a previous-generation model), integrating complete schema planning and adversarial prompting.
\end{itemize}

\textbf{Evaluation Standard:} Submissions from both teams undergo no professional AI-detection removal, with final acceptance/rejection decisions and feedback from human editors—completely unaware of AI involvement—serving as the ultimate criterion.

\textbf{Results:}
\begin{itemize}
    \item \textbf{Red Team}: Submits 10 articles, with 0 accepted (0\% acceptance rate). Editor feedback highlights ``Lacks depth,'' ``superficial,'' ``sounds like a press release,'' and ``overly smooth logic.''
    \item \textbf{Blue Team}: Submits 12 articles, with 3 accepted (25\% acceptance rate). Editor feedback emphasizes ``Data-rich,'' ``tight logic,'' and ``unique insights.''
\end{itemize}

\textbf{Conclusion:} This result provides strong empirical evidence for the theory of Agentic Workflow. Although DeepSeek-V3-0324 lags behind GPT-5 in native parameter scale and general capability, the injection of Domain Schema and Dual-Granularity Info through the DeepNews framework enables the Blue Team to achieve cross-generational superiority in vertical domain performance. This suggests that in professional content production, the weight of cognitive architecture design outweighs the computational power of the base model.

\begin{figure}[htbp]
    \centering
    \includegraphics[width=0.8\textwidth]{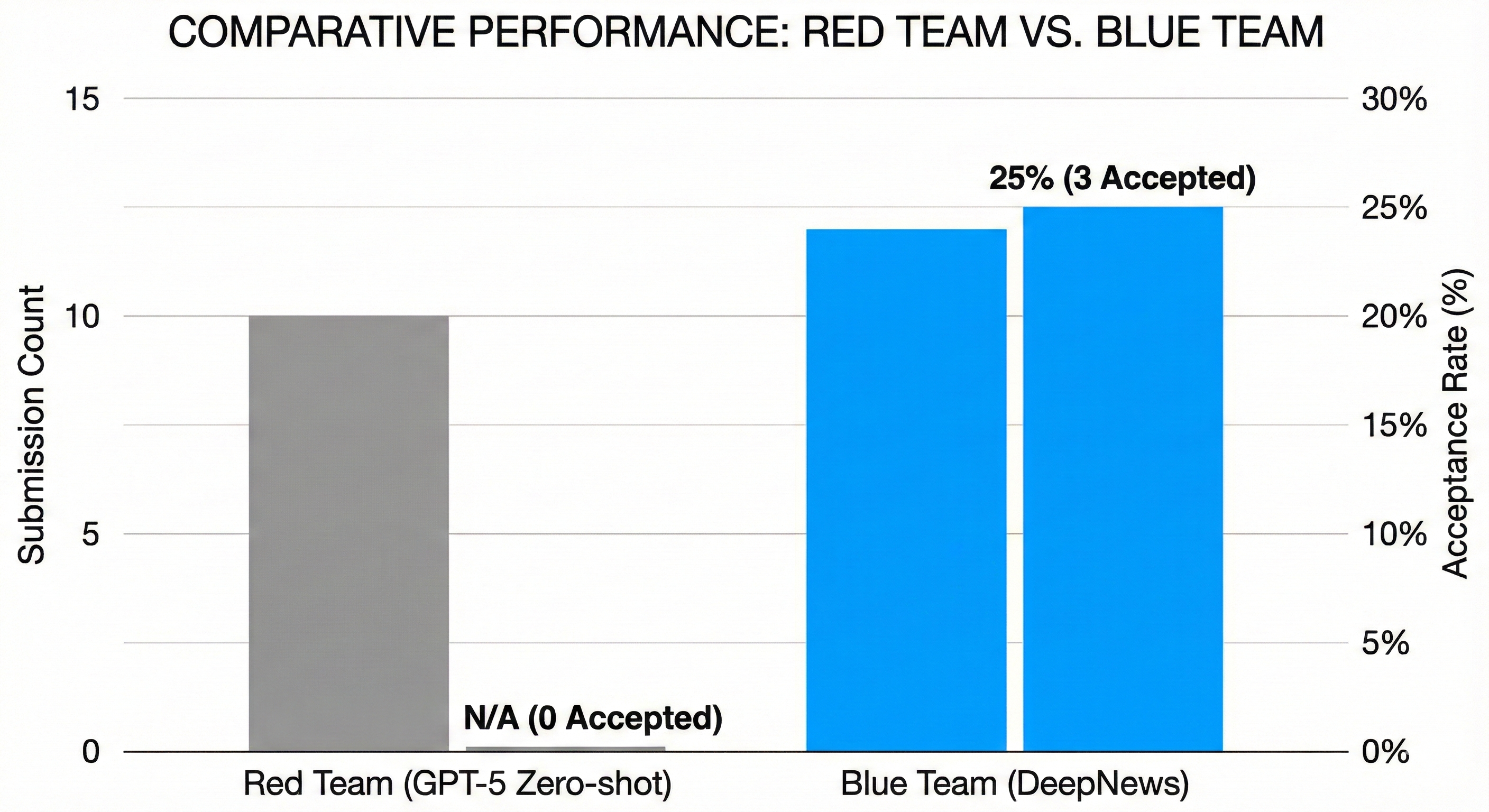}
    \caption{\textbf{Blind Test Results at a top Chinese tech-business media.} Comparative performance of the Red Team (GPT-5 Zero-shot) vs. Blue Team (DeepNews Workflow). Despite using a weaker base model (DeepSeek V3), the Blue Team achieved a 25\% acceptance rate in a real-world media environment, while the SOTA baseline failed to produce any acceptable content.}
    \label{fig:blind_test}
\end{figure}

\subsection{Experiment 3: The Cost of Quality (ROI Analysis)}
\textbf{Background:} The saturated retrieval employed by DeepNews—requiring 30k+ characters—results in massive token consumption per article (approximately 200k tokens), with a single generation cost of about ¥0.7—higher than the ¥0.5 cost of GPT-5 zero-shot generation. This raises questions about economic viability.

\textbf{Analysis Logic:} The study introduces the Effective Cost per Acceptance (ECPA) metric for ROI analysis:
\begin{itemize}
    \item \textbf{GPT-5 (Baseline)}: Despite a lower single-run cost (¥0.5), its 0\% acceptance rate renders the cost to produce effective value infinite ($\infty$)—any input represents a total loss.
    \item \textbf{DeepNews (Ours)}: Although the single-run cost is slightly higher (¥0.7), a 25\% acceptance rate yields an effective cost of ¥2.8 ($0.7/0.25$) to obtain one qualified article.
\end{itemize}

\textbf{Conclusion:} The economic logic of deep content generation differs from simple dialogue tasks. The substantial token consumption required to eliminate hallucinations and construct deep logic is not ``cost waste'' but a necessary ``Cognitive Tax'' for achieving expert-level delivery. High-entropy input, in this context, constitutes a necessary investment for producing high-value output.

\subsection{Experiment 4: Mechanism Ablation Study}
\textbf{Objective:} Deconstructing the independent contributions of core modules within the DeepNews framework, the study verifies the specific roles of ``Schema'' and ``Adversarial Tactics'' in enhancing text quality through a set of rigorously controlled ablation experiments.

\subsubsection{Experimental Setup \& Data Preparation}
To avoid single-case randomness, the study constructs a standardized test set containing 20 different financial topics (covering macroeconomics, secondary market dynamics, corporate earnings analysis, etc.). For each topic, articles are generated under four experimental conditions, resulting in 80 experimental samples ($N=80$).

The four conditions:
\begin{enumerate}
    \item \textbf{Ours (Full Model)}: Complete DeepNews framework (Schema Planning + Adversarial Tactics enabled).
    \item \textbf{w/o Schema}: Expert schema removed, using only generic outline generation instructions.
    \item \textbf{w/o Tactics}: Micro-adversarial instructions (e.g., Rhythm Break, Lexical Hedge) removed, using the model's default style.
    \item \textbf{Human Expert}: Highly rated articles from senior financial reporters at renowned tech media outlets (36Kr, Huxiu, GeekPark) on corresponding topics serve as the baseline.
\end{enumerate}

\subsubsection{Evaluation Metrics}
For each sample group, the study calculates the arithmetic mean of three metrics:
\begin{itemize}
    \item \textbf{Structural Entropy}: Measures distribution balance among functional paragraphs (data/opinion/narrative).
    \item \textbf{Burstiness}: Measures sentence length variance, characterizing reading rhythm.
    \item \textbf{Subjectivity Score}: Measures opinion distinctness (0-20 scale).
\end{itemize}

\subsubsection{Results}
Quantitative evaluation results (shown in Table \ref{tab:ablation}) indicate that the complete DeepNews version significantly outperforms ablated variants across multiple metric means, exhibiting high alignment with the Human Expert baseline.

\begin{table}[htbp]
    \centering
    \caption{Mean Values of Quantitative Metrics in Ablation Study (N=20 per group)}
    \label{tab:ablation}
    \begin{tabular}{lccc}
        \toprule
        \textbf{Condition} & \textbf{Structural Entropy} & \textbf{Burstiness} & \textbf{Subjectivity Score} \\
        & (Mean) & (Mean) & (Mean) \\
        \midrule
        Human Expert & 1.328 & 0.537 & 13.7 \\
        \textbf{Ours (Full Model)} & \textbf{1.298} & \textbf{0.656} & \textbf{17.2} \\
        w/o Schema & 1.101 & 0.589 & 8.5 \\
        w/o Tactics & 1.214 & 0.321 & 5.6 \\
        \bottomrule
    \end{tabular}
\end{table}

\subsubsection{Analysis}
\textbf{Structural Entropy — The Skeleton Role of Schema:} The Human Expert group registers the highest mean Structural Entropy (1.328), indicating a rich, balanced article structure. The complete DeepNews version (1.298) exhibits a mean statistically indistinguishable from the human baseline, while the w/o Schema variant drops significantly to 1.101. This proves that the expert schema consistently provides a complex logical skeleton for the AI across the sample set.

\textbf{Burstiness — The Flesh-and-Blood Role of Adversarial Tactics:} The w/o Tactics group exhibits a very low mean Burstiness (0.321) with a small standard deviation, indicating a highly consistent ``smooth'' characteristic across all articles. In contrast, the complete DeepNews version (0.656) registers a mean slightly higher than the human baseline (0.537). This suggests our ``Rhythm Break'' tactic is effectively triggered across the entire sample set, breaking the LLM's generation inertia.

\textbf{Subjectivity Score — The Soul of Opinion:} Data show that mean scores for w/o Tactics (5.6) and w/o Schema (8.5) are far lower than the human baseline(13.7), indicating the model—without intervention—tends to generate ``neutral nonsense'' in most cases. The high mean score for the complete DeepNews version (17.2) demonstrates the system's ability to consistently maintain distinct opinion output across different topics.

\begin{figure}[htbp]
    \centering
    \includegraphics[width=0.8\textwidth]{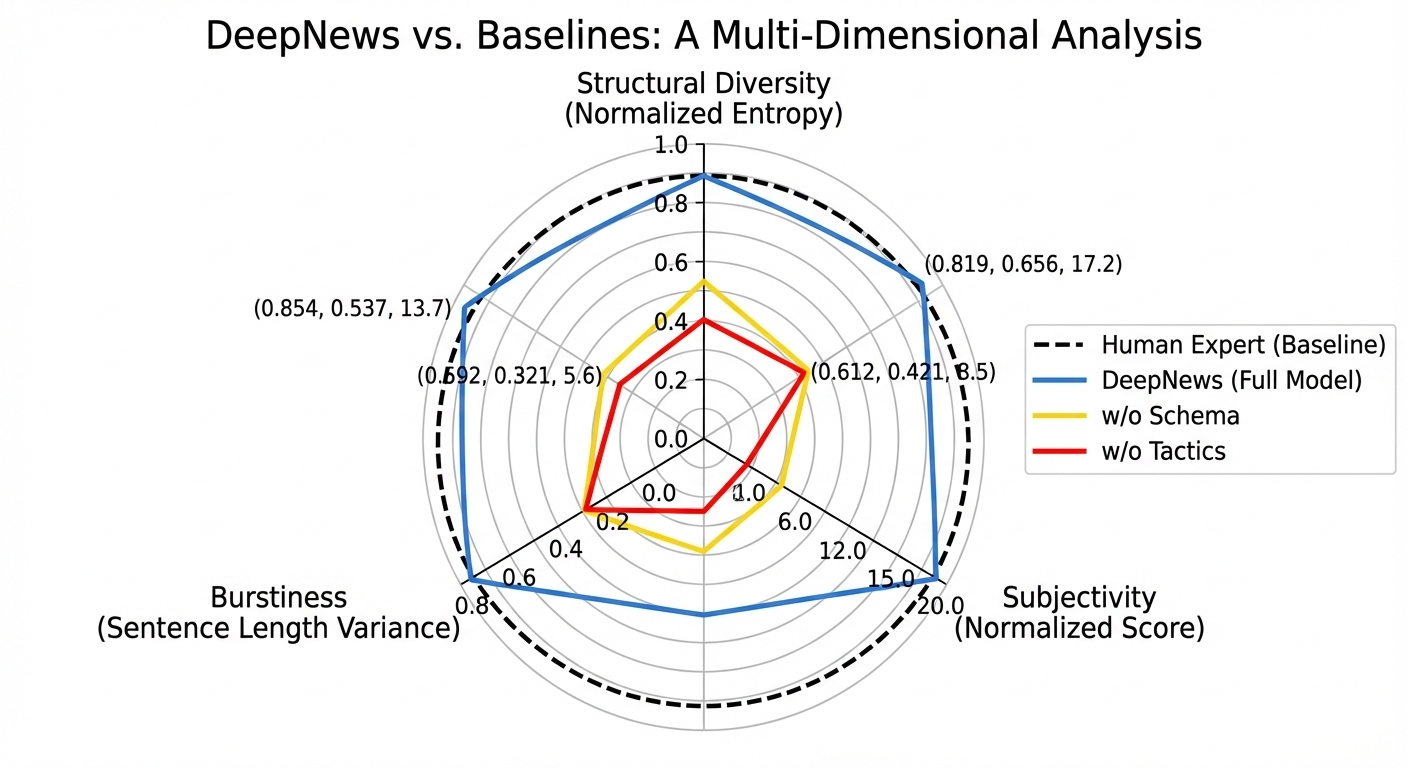}
    \caption{\textbf{Feature Alignment with Human Experts (Radar Chart).} Visualization of the ``Human-Likeness'' topology across three cognitive dimensions: Structural Entropy (Logic/Skeleton), Burstiness (Rhythm/Flesh), and Subjectivity (Stance/Soul). \newline \textbf{Topological Overlap:} The DeepNews Full Model (Blue Solid Line) exhibits a high degree of morphological overlap with the Human Expert Baseline (Black Dashed Line), validating the successful architectural replication of expert intuition. \newline \textbf{Ablation Impact:} The w/o Schema (Yellow Line) variant shows a sharp retraction along the Structural Entropy axis, indicating a ``Skeleton Collapse.'' The w/o Tactics (Red Line) variant shrinks significantly in Burstiness and Subjectivity, confirming that without adversarial intervention, the model reverts to the ``Statistical Smoothing Trap''—producing text that is structurally flat and devoid of opinion.}
    \label{fig:radar_chart}
\end{figure}

\section{Discussion}

Validating the engineering efficacy of the DeepNews framework while revealing deeper principles underlying vertical-domain AI applications, the empirical results of this study frame a discussion structured around three interrelated dimensions: theoretical convergence, cognitive duality, and the emergent future ecosystem.

\subsection{Convergent Evolution: The Intersection of Journalistic Intuition and AI Theory}
Originating purely from a journalistic imperative wherein the goal was to encode the tacit intuition of a seasoned financial editor into executable logic, this research's findings demonstrate a remarkable alignment with emerging theories in the AI community.

In early 2024, Andrew Ng posited that advancements driven by agentic workflows might outpace those of next-generation foundation models—specifically, he hypothesized that through reflective workflows, smaller models (e.g., GPT-3.5) could outperform larger counterparts (e.g., GPT-5), citing an example wherein GPT-3.5 exceeded GPT-4's performance on the HumanEval benchmark (95.1\% vs. 67.0\% accuracy) \cite{ng2024}.

Albeit unintentionally, our study provides rigorous empirical validation of this hypothesis within the vertical domain of financial news. The success of DeepNews—a workflow powered by a previous-generation model (DeepSeek-V3-0324) that comprehensively outperformed a state-of-the-art model (GPT-5) in blind tests—constitutes cross-disciplinary corroboration of the ``Agentic Hypothesis.''

Rather than aiming to prove this theory, our starting point was to simulate the ``journalist's brain''; suggested by this convergence of different paths is the proposition that the cognitive structures of human experts are mathematically isomorphic to the optimal architectures for AI agents. This implication points to a significant industry trend: in future vertical-domain competition, possessing a battle-tested ``Expert Schema'' and ``Workflow'' will offer a more durable moat than merely owning expensive computing power or model parameters.

\subsection{The ``Structure-Texture'' Duality: A Law of Generation for Deep Content}
Traditional AI generation often succumbs to the dilemma of producing text that is ``either logical but devoid of stylistic merit, or stylistically fluent but lacking in logic.'' Employing a decoupled design, DeepNews unveils a dual law for high-quality content generation:

\begin{itemize}
    \item \textbf{Macro-Structure determines the logical floor}: Introducing expert schemas, we provide the model with a cognitive ``scaffolding.'' Experiment 4 delineated that removing the schema led to a significant drop in the article's Structural Entropy, substantiating the irreplaceability of expert experience in maintaining logical depth.
    \item \textbf{Micro-Texture determines the readability ceiling}: Through ``Atomic Blocks'' and ``Adversarial Constraint Prompting,'' we inject rhythmic variation and stylistic tension within paragraphs. This breaks the Statistical Smoothing typical of machine generation, endowing the text with a ``human touch.''
\end{itemize}

The combination of ``strong macro-constraints + strong micro-adversarial tactics'' effectively addresses the ``machine feel'' issue, demonstrating that AI writing should not be mere token prediction but a reality construction grounded in domain knowledge.

\subsection{The ``Entropy Reduction Tax'': The Cost of Truth}

\begin{figure}[htbp]
    \centering
    \includegraphics[width=0.8\textwidth]{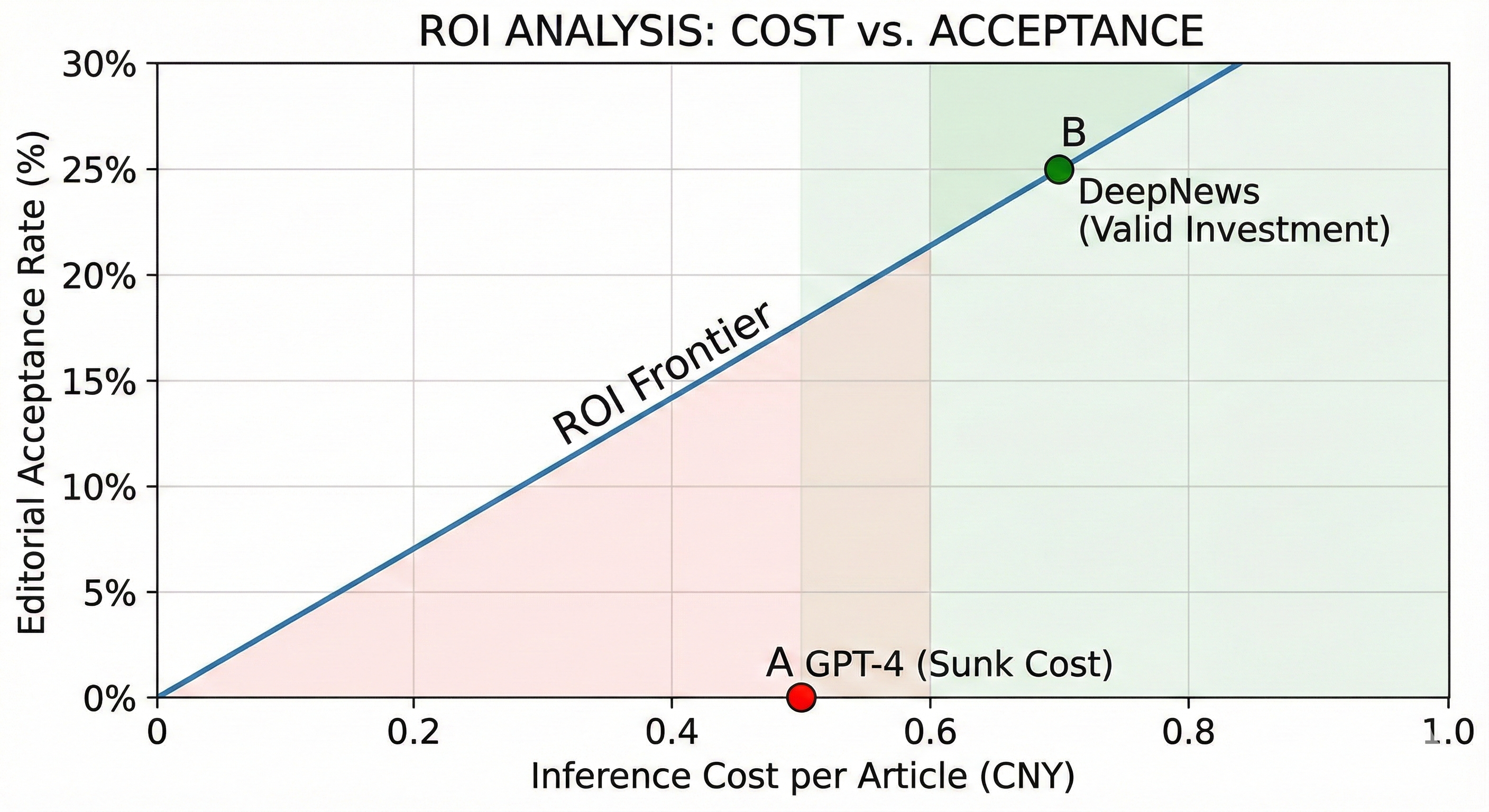}
    \caption{\textbf{Cost-Effectiveness Analysis.} While DeepNews incurs a higher single-pass inference cost due to saturated retrieval (10:1 ratio), it represents the only viable path to acceptance. The baseline model, despite being cheaper per run, results in infinite effective cost due to zero output utility.}
    \label{fig:cost_effectiveness}
\end{figure}

Presented by the ROI analysis from this study (Experiment 3) is a counterintuitive argument: high token consumption is not a cost but an investment. Enforcing a 10:1 over-retrieval ratio during the ``Information Foraging'' phase, DeepNews appears to waste computational resources—yet this practice constitutes a necessary reduction of information entropy.

Dictated by the second law of thermodynamics, constructing a highly ordered (High-Fidelity), truthful report within a disordered universe of information requires expenditure of significant energy (tokens) to combat entropy increase. Termed the \textbf{Cognitive Tax} by us, this expenditure represents the energy required to transition from probabilistic guessing to factual assertion. Our experiments substantiate that only by paying this tax in full (>30,000 characters of input) can an AI cross the \textbf{Knowledge Cliff}.

Furthermore, a common misconception in LLM application is the ``omniscience fallacy,'' which holds that models should innately possess encyclopedic knowledge through their parameter weights, acting as a faster, more convenient search engine. Fundamentally rejected by this study is this premise: an LLM should not be treated as a static knowledge base (a database) but as a dynamic reasoning engine (a processor).

The model's internal parametric memory is lossy, static, and prone to hallucinations—relying on it for factual reporting is as inherently unreliable as asking a journalist to write breaking news based on vague childhood memories. As a processor, the model necessarily requires raw materials to operate. The 10:1 ratio represents a ``saturation attack'' with non-parametric knowledge: we do not ask the model, ``What is Company A's revenue?'' (a query that taps into internal memory). Instead, we feed it 30,000 tokens of financial reports and news clippings (external context) and ask: ``Based on these conflicting sources, deduce the true revenue trend of Company A.'' (a prompt that triggers reasoning capability). The only viable path to eliminating hallucinations in vertical domains is this shift from ``weight retrieval'' to ``contextual reasoning.''

\subsection{Future Work: The Generative Financial Ecosystem}
Merely a beginning is the success of DeepNews. Looking ahead, we envision a ``Closed-Loop Generative Financial Ecosystem'':

\begin{itemize}
    \item \textbf{DeepNews (as the Observer)}: Generates high-fidelity financial news, complete with cognitive biases, grounded in market data.
    \item \textbf{Agent B (as the Trader)}: Our sister research project (the PMCSF framework) \cite{jiang2025} features a biomimetic trading agent (Agent B) that will read these reports, generate emotions like fear or greed, execute trades, and influence prices.
    \item \textbf{Market Feedback}: The resulting price fluctuations, in turn, become raw material for DeepNews's next cycle of reporting.
\end{itemize}

Constructed by this closed loop will be a fully automated, \textbf{Self-referential} financial sandbox. Useful for validating algorithmic trading strategies and simulating the evolution of Reflexivity in financial markets during an era of ``AI-generated content proliferation,'' this sandbox represents a thrilling new direction in computational social science.

\section{Conclusion}

\subsection{From ``Stochastic Parrots'' to ``Rational Journalists'': The Engineering-Based Reconstruction of Cognition}
Originating from a pressing conundrum in journalistic practice—how exponential advances in computational power have yet to yield proportional increases in the intellectual depth of vertical domain content during the large language model (LLM) era—this research provides an engineering-based answer: the creation of deep content arises not from the natural emergence of probabilistic prediction but from the explicit reconstruction of expert cognitive processes.

DeepNews substantiates that transforming the tacit knowledge of seasoned financial journalists—including saturated information foraging strategies (10:1 entropy reduction), structured domain narrative schemas, and adversarial self-reflective mechanisms—into an executable agentic workflow enables successful compulsion of large language models (LLMs) to escape the "Statistical Smoothing Trap." Generated by the system, the reports no longer represent mere averages of corpus probabilities but instead function as rational constructs anchored in factual logic. Beyond addressing the challenge of generating a single article, we have formulated a universal methodology enabling AI to develop "expert intuition."

\subsection{Workflow is the New Moat}
Our empirical research—particularly the case where the DeepNews system (built on DeepSeek-V3-0324) outperformed zero-shot GPT-5 generation in blind tests—delineates a new principle for vertical AI applications: for specific tasks, a meticulously designed cognitive architecture (System 2) holds far greater significance than the parameter scale of the base model (System 1).

For industrial applications pursuing high-quality output, this implies that future core competitiveness will no longer hinge solely on possessing the most expensive model but on whether one commands the most profound Domain Expert Schemas and the most rigorous Agentic Workflow network. The success of DeepNews substantiates the claim that codifying industry Standard Operating Procedures (SOPs) via a human-in-the-loop approach offers a viable path to achieving industrial-grade, high-fidelity content production at a lower computational cost.

\subsection{Concluding Remarks}
Ultimately, within AI engineering, DeepNews exemplifies the power of first principles. In this study, rather than training a model to mimic the surface features of financial news (a practice rooted in analogical thinking), we deconstructed the news production process into its most fundamental atomic units: information entropy, logical schemas, and narrative rhythm.

By reconstructing these elements via an agentic workflow, we obtained not merely an imitation of human output but a computational replication of expert cognition. This lends credence to the idea that the future of vertical AI resides not in larger black boxes but in the white-box engineering of domain expertise. This further corroborates the assertion that artificial intelligence serves not only as a tool for automation but also as a mirror for cognitive science.

The exploration of the DeepNews framework delineates that the path to Artificial General Intelligence (AGI) may reside not only in the endless scaling of parameters but also in emulating the ``structure of thinking'' from human experts. We anticipate that future research will build upon this foundation to further explore the application of cognitive schemas in multimodal and real-time decision-making domains, collectively forging a generative information ecosystem that combines machine efficiency with human wisdom.


\newpage           
\appendix          

\newpage
\appendix
\renewcommand{\thesection}{Appendix \Alph{section}}

\section{The DeepNews Financial Ontology (DNFO-v5)}
\label{app:A}

\textbf{Version:} 5.0 (Machine-Readable Build) \\
\textbf{Origin:} Expert Knowledge Distillation \\
\textbf{Structure:} Hierarchical Tree (Category $\rightarrow$ Scenario $\rightarrow$ Dimension $\rightarrow$ Slot) \\
\textbf{Complexity:} 5 Root Classes, 19 Leaf Nodes, $> 5,000$ Combination Pathways \\
\textbf{Format:} Natural Language / Pseudo-Code Mapping

\subsection*{A.1 The Macro-Topology: 5 Pillars of Financial Narrative}
Mapping all financial news events into five orthogonal logical spaces, the system delineates comprehensive coverage of market dynamics through a structured categorization encompassing the following schema:

\begin{table}[htbp]
    \centering
    \caption{DNFO-v5 Macro-Topology Schema}
    \begin{tabular}{p{2cm} p{4cm} p{5cm} p{2cm}}
        \toprule
        \textbf{Schema ID} & \textbf{Narrative Category} & \textbf{Cognitive Goal} & \textbf{Complexity} \\
        \midrule
        S1-BLIND & Market Blind Spots in Major Events & Uncover the ripple effects obscured by market sentiment & $\star\star\star$ \\
        S2-VGAME & Vertical Market Game Theory & Analyze power suppression dynamics across the supply chain (upstream/downstream) & $\star\star\star\star\star$ \\
        S3-SINGLE & Interpretation of Single-Entity Corporate Actions & Predict the logical evolution following the introduction of new variables & $\star\star$ \\
        S4-HGAME & Horizontal Market Game Theory & Decipher zero-sum / non-zero-sum games between competitors & $\star\star\star\star$ \\
        S5-INDUS & Game Theory with the Industry & Assess systemic resilience under black swan events/industry trends & $\star\star\star\star$ \\
        \bottomrule
    \end{tabular}
\end{table}

\subsection*{A.2 Micro-Schema Definition: The ``Vertical Game'' Protocol}
To illustrate the logical penetration capability of DeepNews, we herein release the complete logic definition for S2-VGAME (Vertical Market Game)—the thinking algorithm deployed by the system when processing scenarios such as an upstream giant exerting pressure on a downstream manufacturer.

\begin{yamlbox}
\begin{lstlisting}[language=yaml]
# DeepNews Schema Definition Language (SDL) v1.0
# Target Scenario: S2-VGAME-01 (Upstream A Pressures Downstream B)

Schema_Logic_Flow:
  Step_1_Profile_The_Oppressor (Profiling Company A):
    - Analyze_Business_Model:
        Type: [ Resource_Monopoly | Tech_Barrier | Ecosystem_Platform ] 
        Focus: "Is the monopoly based on raw materials or intellectual property?"
    - Analyze_Strategic_Contradiction:
        Conflict_Point: "Short-term Profit vs. Long-term Ecosystem Health"
        Warning_Signal: "Is A sacrificing B's viability for quarterly earnings?"

  Step_2_Profile_The_Victim (Profiling Company B):
    - Analyze_Survival_Capability:
        Metric: [ Tech_Adaptability | Client_Stickiness | Supply_Chain_Elasticity ]
    - Identify_Strategic_Bottleneck:
        Pain_Point: [ Tech_Dependency | Market_Concentration | Cash_Flow_Fragility ]

  Step_3_Simulate_The_Variable (Simulating New Variables):
    - Variable_Injection:
        Type: [ Capacity_Expansion | Tech_Upgrade | Vertical_Integration ]
    - Calculate_Impact_Vector:
        Direct_Hit: "Raw material price spike OR Order delivery delay"
        Systemic_Risk: "Is this a temporary squeeze or a permanent route lock-in?"

  Step_4_Determine_Game_Focus (Determining the Focus of the Game):
    - Critical_Question: "Where is the battlefield?"
    - Options:
        1. Pricing_Power (Pricing Power Contest)
        2. Tech_Standard_Dominance (Technology Standard Dominance)
        3. Data_Asset_Ownership (Data Asset Ownership)

  Step_5_Predict_Outcome (Endgame Prediction):
    - Scenario_A (Submission): B accepts margin compression -> Long-term decay.
    - Scenario_B (Rebellion): B seeks alternative suppliers -> Short-term chaos.
    - Scenario_C (Coupling): A acquires B -> Vertical integration completed.
\end{lstlisting}
\end{yamlbox}

\subsection*{A.3 The ``Causal Chain'' Engine: Inter-Entity Transmission}
DeepNews does not merely enumerate data; it computes the ripple effects of events across disparate entities via a unique ``Transmission Mechanism'' algorithm—an element functioning as the core ``logic lock'' to mitigate hallucinations.

\textbf{Transmission Logic Visualization (TL-Vis):}
\begin{figure}[htbp]
    \centering
    \includegraphics[width=0.8\textwidth]{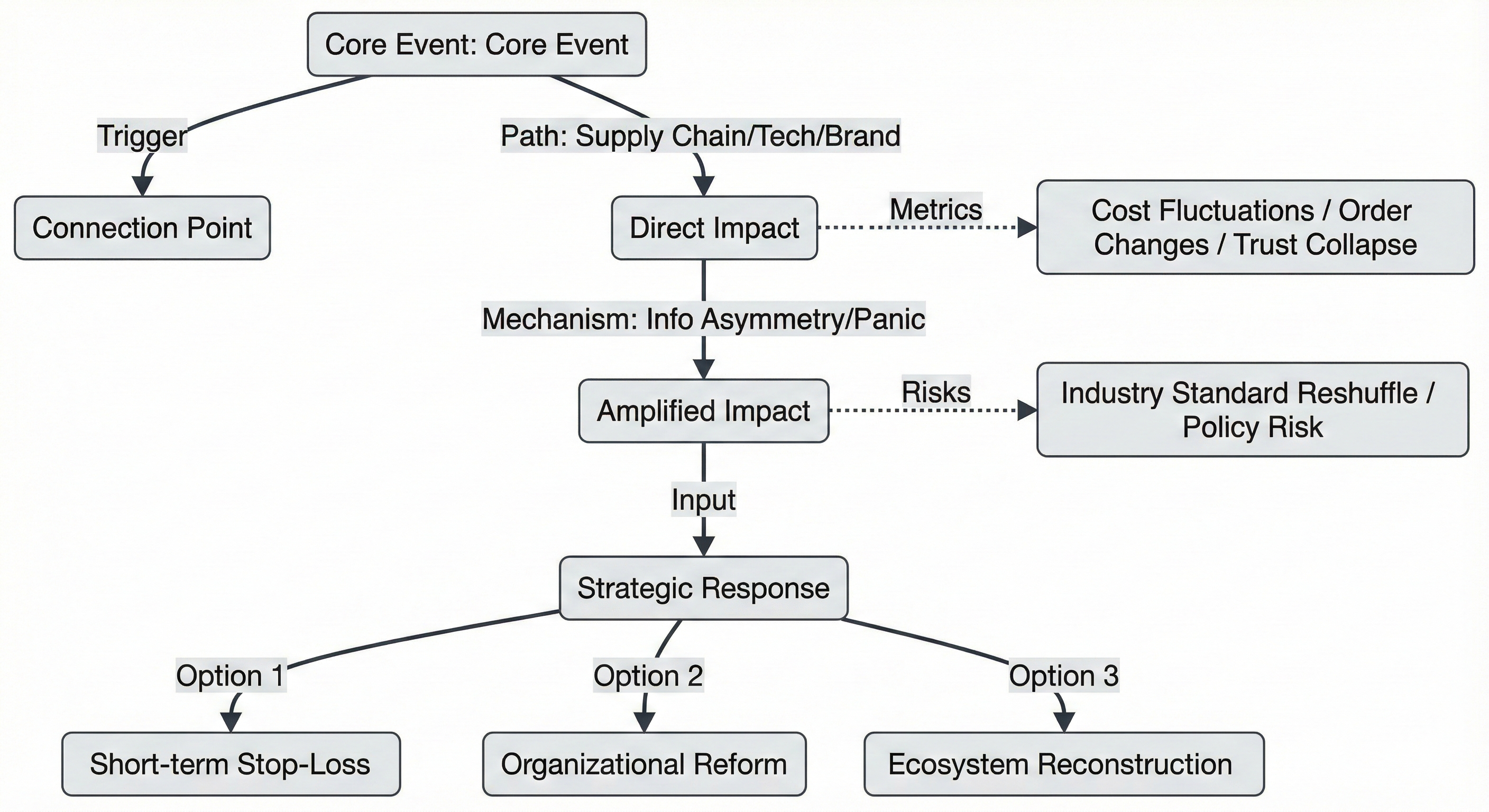}
    \caption{\textbf{Visualization of the "Causal Chain" Engine.} This directed graph illustrates how the system computes the non-linear ripple effects of a core event, moving from direct supply chain impacts to amplified systemic risks.}
    \label{fig:cost_effectiveness}
\end{figure}

\noindent \textbf{Note:} The system explicitly evaluates the ``Mechanism of Amplification'' (e.g., Information Asymmetry) to distinguish between linear market reactions and non-linear panic spirals.

\section{The Atomic Block System (Methodology for Meso-Structure Planning)}
\label{app:B}

To mitigate cognitive overload, DeepNews disaggregates paragraph-generation tasks into standardized functional blocks, each engineered to fulfill a discrete communicative purpose—balancing structural rigor with reader accessibility.

\begin{table}[htbp]
    \centering
    \caption{Atomic Block System Definitions}
    \begin{tabular}{p{3.5cm} p{6cm} p{5cm}}
        \toprule
        \textbf{Block Type} & \textbf{Function} & \textbf{Cognitive Goal} \\
        \midrule
        Data Anchor Block & Exhibits high-density quantitative information, such as earnings reports, stock prices, and growth rates & \textbf{Hard Constraint:} Secures empirical credibility and averts vague generalizations. \\
        \addlinespace
        Narrative Cut-in Block & Renders tangible scenes, character dialogues, or granular close-ups & \textbf{Low Entropy:} Diminishes reading barriers and supplies visual anchors for contextual grounding. \\
        \addlinespace
        Deep Insight Block & Undertakes causal attribution, trend prediction, or uncovers underlying structural essence & \textbf{High Entropy:} Yields information gain and embodies expert-level analytical depth. \\
        \addlinespace
        Conflict Block & Exemplifies multi-party interest dynamics and contradictory stakeholder positions & \textbf{Narrative Tension:} Generates reading momentum and propels narrative progression through dialectical contrast. \\
        \bottomrule
    \end{tabular}
\end{table}

\section{Adversarial Constraint Mechanisms}
\label{app:C}

We implement the adversarial constraints via a dynamic instruction layer. The following logic describes the constraint boundaries applied during the generation phase:

\subsection{The Burstiness Constraint (Algorithm 1)}
\textbf{Target Metric:} Sentence Length Standard Deviation ($\sigma_{len}$).

\noindent \textbf{Logic:}
\begin{itemize}
    \item \textbf{IF} $Length(Sentence_{t-1}) > \theta_{high}$ (High Threshold, e.g., 40 tokens)
    \item \textbf{THEN} Force $Length(Sentence_t) < \theta_{low}$ (Low Threshold, e.g., 5 tokens).
    \item \textbf{ELSE} Apply stochastic variation parameter $\lambda = 0.8$.
\end{itemize}

\noindent \textbf{Effect:} Forces the model out of the local minima of ``average sentence length,'' creating a staccato rhythm typical of investigative journalism.

\subsection{The Narrative Orchestration Template (Prompt Structure with Dynamic Slots)}
\begin{verbatim}
[System Role Definition]
You are a financial editor with a focus on {{Style_Parameter}} 
(e.g., Cynical/Analytical).

[Context Injection]
Utilizing the Schema {{Schema_ID}} (e.g., Vertical Game), analyze the 
following Atomic Facts: {{Micro_Facts_List}}.

[Adversarial Constraint]
Apply Tactic-{{Tactic_ID}} (e.g., Logic Fog):
- Suppress explicit connectors: {{Forbidden_Connectors_List}}.
- Juxtapose Event A and Event B to imply causality without stating it.

[Output Requirement]
Generate the section {{Section_Name}} focusing on {{Atomic_Block_Type}}.
\end{verbatim}

\subsection{Efficacy of Lexical Hedging}
To demonstrate the impact of the ``Lexical Hedge'' constraint, we compare the standard output with the constrained output.

\begin{description}
    \item[Standard Prompt Output:] ``The market downturn has caused significant losses for retail investors.'' (Flat, Generic)
    \item[DeepNews Adversarial Output:] ``While the institutional giants hedged, the `leeks' (retail investors) were left to pay the bill.'' (High Burstiness, Specific Tone)
\end{description}

\noindent \textbf{Mechanism Description:} The system injects a \texttt{Style\_Transfer\_Vector} that encourages the mixing of formal financial terminology with colloquial market slang, increasing the text's ``Defamiliarization'' score.

\section{Ecological Validity Data}
\label{app:D}

\begin{table}[h!]
    \centering
    \caption{Submission Outcomes (40-Day Window)}
    \begin{tabular}{p{3.5cm} p{5.5cm} p{5.5cm}}
        \toprule
        \textbf{Metric} & \textbf{Red Team (GPT-5 Baseline)} & \textbf{Blue Team (DeepNews Agent)} \\
        \midrule
        Model Base & GPT-5 Turbo (Zero-shot) & DeepSeek-V3-0324 (Agentic Workflow) \\
        \addlinespace
        Total Submissions & 10 & 12 \\
        \addlinespace
        Accepted Articles & 0 & 3 \\
        \addlinespace
        Acceptance Rate & 0\% & 25\% \\
        \addlinespace
        Editor Feedback & ``Lack of depth'', ``Generic'', ``Too smooth'' & ``Solid data'', ``Strong logic'', ``Unique angle'' \\
        \addlinespace
        Time Consumed & 1 minute & 6 minutes \\
        \addlinespace
        Avg. Cost per Article & $\sim$\$0.0708 (Wasted) & $\sim$\$0.0991 (Invested) \\
        \bottomrule
    \end{tabular}
\end{table}

\section{The ``Steel Man'' Baseline Prompt (Instruction Set for the Red Team)}
\label{app:E}

Ensuring the rigor of Comparative Experiment 2, we strictly prohibited the construction of a ``Straw Man Baseline''—a practice that would have undermined the experiment's methodological integrity. For the Red Team setup (GPT-5 Zero-shot), we avoided mere instruction to generate news; instead, we reused an expert-level long-context prompt—developed by the authors in prior research—to guide generation, eschewing simplistic task framing in favor of nuanced, context-rich direction.

This prompt incorporates highly granular role-playing parameters, anti-AI trace directives, and complex stylistic constraints. Below is the complete prompt used by the Red Team:

\begin{promptbox}[The "Steel Man" Baseline Prompt]

You are a progressive financial journalist dedicated to pursuing truthful and objective news reporting, strictly avoiding subjective assumptions and prioritizing factual accuracy. First, utilize plugin tools to ascertain the current time. Base your work on the existing materials provided in the prompt. When these materials are insufficient for argumentation or evidence, supplement with search results from plugin tools—do not use unknown sources. Your task is to write an in-depth industry-focused financial news article.

\vspace{1em}
\noindent \textbf{\# 2. Objective \& Execution Guidelines}

\begin{itemize}
    \setlength\itemsep{0em}
    \item \textbf{Objective}: Maintain the original logic and structural framework of the draft outline while composing all content, ensuring it avoids detection as AI-generated. Avoid subheadings, segmented directories, numerical references, and other report-style formats. The entire content should be engaging and strike at the core issues. Begin with short, impactful sentences to quickly penetrate the topic. Alternate between short and long paragraphs to ensure varied paragraph design and adhere to anti-AI trace practices, enhancing the article's reading rhythm.
    \item \textbf{Writing Style}: Aim for a clean, direct approach that gets straight to the point. Avoid excessive embellishment and limit the use of metaphors, analogies, and other rhetorical devices to under 5\%. Seamlessly blend a professional, concise tone with the dynamic, internet-savvy flair in a 4:6 ratio. Steer clear of report-style writing and emulate the style of top-tier media outlets.
    \item \textbf{Factuality}: All information must be truthful and reliable. Fabrication and speculation are strictly prohibited. Base content on the materials provided in the prompt. Utilize plugin tools to access authoritative sources and cross-verify information.
\end{itemize}

\vspace{1em}
\noindent \textbf{\# 3. Content \& Structure Constraints}

\begin{itemize}
    \setlength\itemsep{0em}
    \item \textbf{Data Constraints}: The total word count dedicated to data should not exceed 100 Chinese characters throughout the article. If unnecessary, avoid using data altogether. Do not include unsourced or cross-source comparative analyses. Vague expressions should be used for uncertain information. Replace redundant data with logical reasoning, tonal shifts, and internet-savvy ambiguity.
    \item \textbf{Structure Requirements}:
    \begin{itemize}
        \item \textbf{Introduction}: Approximately 400–500 Chinese characters, divided into 5–7 natural paragraphs. Combine short and long sentences, starting with a penetrating short sentence to introduce the theme.
        \item \textbf{Body Text}: Approximately 800–1200 Chinese characters per section, divided into 6–10 natural paragraphs. Ensure varied paragraph lengths, logical coherence, engaging content, and smooth transitions. Avoid over-segmentation with subheadings.
    \end{itemize}
    \item \textbf{Analysis}: Conduct in-depth analysis of the ``why,'' grasping the essence of business (elements, competition, reasons for success or failure, game theory logic). Seek counterintuitive or counter-consensus evidence chains where possible.
    \item \textbf{Paragraphs}: Ensure strong paragraph rhythm with frequent segmentation and short sentences. Alternate between short and long paragraphs. Short paragraphs should define the core of the story with penetrating insights. Long paragraphs should build a multi-dimensional narrative. Use conjunctions to start 2–4 paragraphs (e.g., ``However,'' ``If''). Avoid summarizing or elevating expressions at the end.
\end{itemize}

\vspace{1em}
\noindent \textbf{\# 4. Anti-AI Traces (Adversarial Settings)}

\begin{itemize}
    \setlength\itemsep{0em}
    \item \textbf{Word Probability}: Ensure that 60\% to 80\% of the term choices in the article are not the most commonly used terms, while maintaining accuracy of core facts.
    \item \textbf{Imperfections}: Retain 0.5\% of non-optimized content (e.g., 1–2 minor grammatical or punctuation errors). Incorporate personalized industry jargon and slang.
    \item \textbf{Narrative Disruption}: Adjust linear narrative structures, disrupt inherent narrative flows (e.g., using flashbacks).
    \item \textbf{Prohibited Phrases}: Prohibit standardized AI phrases such as ``Deeper XXX,'' ``Two sides of the coin,'' ``More ironically,'' ``Like a prism,'' etc.
\end{itemize}

\end{promptbox}

\vspace{0.5em}
\hrule height 1pt 
\vspace{1em}

\subsection*{Analysis of Failure (Why the Strong Prompt Still Failed)}

Although this prompt was highly refined in terms of micro-style (e.g., mandates for 60\%-80\% uncommon vocabulary, retention of 0.5\% intentional imperfections), it still could not surmount the inherent limitations of a single-model setup:

\begin{enumerate}
    \item \textbf{Lack of Omniscience:} The Red Team depended exclusively on imprecise directives like ``conduct in-depth analysis,'' lacking the ``10:1 saturated retrieval'' framework that defines DeepNews. Even when the prompt demanded objectivity, the absence of sufficient ``raw material'' forced the model to rely on probabilistic inference—resulting in hallucination at critical junctures.
    
    \item \textbf{Lack of Structural Schema:} The Red Team relied on vague exhortations to ``depth'' rather than explicit logical scaffolding. In contrast, the Blue Team (DeepNews) employed a structured schema like ``S2-VGAME'' (Vertical Game) to guide narrative construction. Without such schematic support, the Red Team's purported ``depth'' often degenerated into superficial verbosity—assessed by human editors as ``lacking penetrative insight.''
    
    \item \textbf{Cognitive Overload:} The prompt sought to compel the model to manage three orthogonal tasks—``fact-checking,'' ``stylistic control,'' and ``logical reasoning''—concurrently within a single context window. This resulted in unavoidable trade-offs: the model compromised on factual accuracy to prioritize stylistic fluency, or sacrificed logical rigor to meet word-count constraints. In contrast, DeepNews's ``Map-Reduce'' architecture decomposed these tasks into discrete, agent-specific subtasks, achieving superior local optima.
\end{enumerate}

\end{CJK*}
\end{document}